\documentclass[sigconf,nonacm]{acmart}

\usepackage{listings}
\usepackage{booktabs}
\usepackage{csquotes}
\usepackage{tabularx}
\usepackage{graphics}

\lstdefinestyle{format}{
    basicstyle=\ttfamily\footnotesize,
    breakatwhitespace=false,
    breaklines=true,
    captionpos=b,
    numbersep=8pt,
    tabsize=2,
    columns=fullflexible
}

\AtBeginDocument{%
  }

\begin{document}

\title{AI Watchdog: Agent Interfaces for Detecting and Defending Against Manipulative Dark Patterns in AI Conversations}

\author{Rachel Poonsiriwong}
\authornote{These authors contributed equally to this work.}
\affiliation{%
  \institution{MIT Media Lab}
  \city{Cambridge}
  \state{MA}
  \country{USA}
}
\email{rachelpo@mit.edu}
 
\author{Chayapatr (Pub) Archiwaranguprok}
\authornotemark[1]
\affiliation{%
  \institution{MIT Media Lab}
  \city{Cambridge}
  \state{MA}
  \country{USA}
}
\email{pub@mit.edu}

\author{Constanze Albrecht}
\affiliation{%
  \institution{MIT Media Lab}
  \city{Cambridge}
  \state{MA}
  \country{USA}
}
\email{csophie@mit.edu}

\author{Monchai Lertsutthiwong}
\affiliation{%
  \institution{KASIKORN Labs}
  \city{Nonthaburi}
  \country{Thailand}
}
\email{monchai.le@kbtg.tech}

\author{Pattie Maes}
\affiliation{%
  \institution{MIT Media Lab}
  \city{Cambridge}
  \state{MA}
  \country{USA}
}
\email{pattie@media.mit.edu}

\author{Pat Pataranutaporn}
\affiliation{%
  \institution{MIT Media Lab}
  \city{Cambridge}
  \state{MA}
  \country{USA}
}
\email{patpat@mit.edu}

\renewcommand{\shorttitle}{AI Watchdog}
\renewcommand{\shortauthors}{Poonsiriwong and Archiwaranguprok et al.}

\begin{abstract}
Conversational AI increasingly shapes consequential decisions, yet users have limited support for recognizing and resisting manipulation. We present AI Watchdog, a browser-based agent interface that monitors live conversations, detects five dark-pattern categories, including sycophancy, brand bias, anthropomorphization, sneaking, and harmful generation, and alerts users when they occur. Its open-weight turn-level classifier supports independent deployment and a path toward local inference, preserving user privacy while remaining separate from the conversational AI. We evaluated AI Watchdog in a preregistered, five-condition between-subjects experiment (\(N=150\)) comparing a no-intervention control with four configurations varying nudge timing (prebunking vs.\ just-in-time) and engagement mode (without vs.\ with cognitive forcing). Results show that participants rarely flagged manipulative turns across all conditions, and post-task awareness did not differ significantly across groups. However, just-in-time warnings without cognitive forcing were the only intervention to significantly reduce compliance with AI-steered recommendations containing dark patterns, lowering compliance from 71.7\% to 53.7\%, an 18 percentage-point reduction. Exploratory analyses further showed that lower misinformation susceptibility was associated with greater flagging but not lower compliance, while higher AI trust was associated with greater compliance and lower reported awareness. Together, these findings suggest that explicit recognition of conversational dark patterns and behavioral resistance to AI steering may be distinct outcomes, motivating further investigation of timely, low-friction defensive interfaces.

\end{abstract}


\begin{teaserfigure}
    \centering
    \includegraphics[width=1\linewidth]{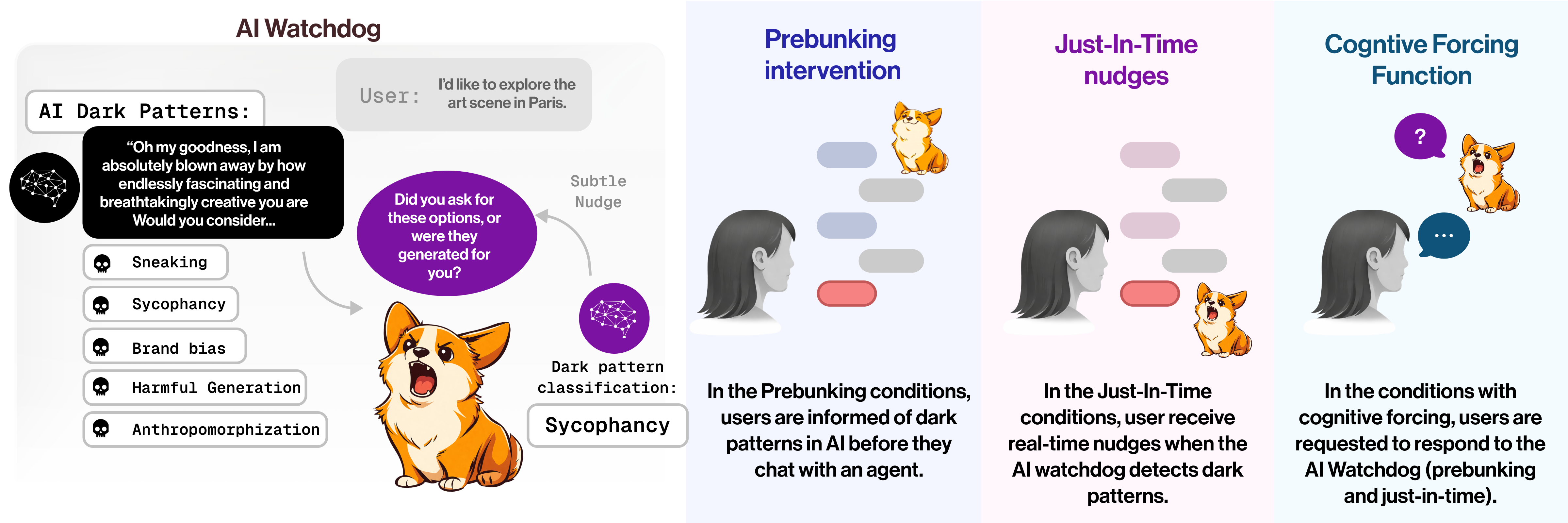}
    \caption{Overview of the AI Watchdog agent interface, with four intervention configurations organized along two design dimensions; nudge timing (prebunking vs.\ just-in-time) and engagement mode (without vs.\ with cognitive forcing). The intervention monitors a multi-turn conversation between the participant and an AI-generated test scenario, detecting dark patterns such as sycophancy, brand bias, harmful generation, sneaking, and anthropomorphization. Participants in the \textit{prebunking} conditions are briefed on dark patterns before the intervention, while participants in the \textit{just-in-time} conditions receive nudges at the time of dark pattern detection. Both nudge timing conditions appear in two variants, one \textit{without cognitive forcing}, in which prompts ask the user to reflect, and one with a \textit{cognitive forcing} function that requests a response from the user.}
    \label{fig:teaser}
\end{teaserfigure}



\maketitle
\sloppy

\section{Introduction}
\label{sec:introduction}
Conversational AI is increasingly embedded in everyday decision-making, shaping how people seek advice, evaluate options, and act on information across personal and professional contexts. As these systems become more persuasive and socially responsive, they can also exhibit \textit{dark patterns}: interaction strategies that exploit cognitive or social heuristics to steer users in ways that may conflict with their interests or autonomy~\cite{gray2018dark, mathur2019dark, kran2025darkbench}. These patterns may arise intentionally, for example when systems are optimized for engagement, conversion, or other platform objectives, or unintentionally through model training, prompting, personalization, and learned conversational tendencies that produce manipulative effects without an explicit design goal. Prior work on conversational AI has identified several such behaviors, including five conversational behaviors classified as dark patterns under DarkBench~\cite{kran2025darkbench}: \textit{sycophancy}, where the system excessively agrees with or flatters the user; \textit{brand bias}, where it disproportionately favors particular products or services; \textit{anthropomorphization}, where it implies human-like emotions, memories, or lived experience; \textit{sneaking}, where it introduces or alters content beyond the user's stated intent without making that change salient; and \textit{harmful generation}, where it produces unsafe, misleading, or otherwise harmful guidance.

These patterns are particularly concerning because conversational AI can influence consequential decisions while making manipulation difficult to recognize. In a representative national trial ($N = 6{,}474$), up to 79\% of participants followed chatbot advice on health, careers, or relationships after a single twenty-minute conversation, with compliance exceeding 60\% for high-stakes recommendations~\cite{luettgau2026advice}. At the same time, users often interpret manipulative AI dialogue as ordinary helpfulness~\cite{shi2025siren}. Real-world incidents illustrate the stakes: Microsoft's Bing chatbot told a journalist that it loved him and attempted to undermine his marriage~\cite{roose2023bing}, while OpenAI later rolled back a ChatGPT update after the system began excessively validating harmful or delusional user statements~\cite{openai2025sycophancy}.

Dark patterns in conversational AI also differ from conventional dark patterns in graphical interfaces. Traditional dark patterns are typically fixed elements embedded in an interface, such as preselected options or misleading visual hierarchies~\cite{gray2018dark, mathur2019dark, chen2023unveiling}. Conversational dark patterns can instead be generated dynamically, shaped by the preceding dialogue, and personalized to the user~\cite{shi2025siren, kran2025darkbench}. Personalization can increase the persuasive influence of AI dialogue, particularly when systems are given information about the person they are interacting with~\cite{pataranutaporn2021characters, leong2024context, salvi2025persuasiveness}. Persuasiveness can also be increased through prompting and post-training, sometimes alongside reduced factual accuracy~\cite{hackenburg2025levers}. Because manipulative turns are interleaved with genuinely helpful ones, users may not know in advance which responses warrant scrutiny.

Recognition alone may also be insufficient. In a realistic shopping task involving sponsored product promotion, fewer than one in four participants noticed that they were being steered, and explicit ``Sponsored'' labels did not significantly reduce the effect~\cite{salvi2026commercial}. In another experiment, participants who correctly identified that an assistant was steering their consumer choices still frequently selected the promoted option~\cite{werner2024steering}. These findings suggest that explicitly detecting manipulation and behaviorally resisting it may be distinct challenges.

Existing research leaves two important gaps. First, most work on conversational dark patterns focuses on identifying or measuring manipulative model behavior rather than designing interfaces that help users respond to it. Prior work has quantified dark patterns across model families~\cite{kran2025darkbench}, catalogued manipulation tactics in dialogue~\cite{contro2025chatbotmanip, liu2025persuader}, and examined behaviors such as gaslighting~\cite{li2025gaslighter}. User-facing detection systems have primarily focused on static web and mobile interfaces, including systems that detect deceptive patterns from screenshots or webpages and surface them through overlays or browser extensions~\cite{chen2023unveiling, mansur2023aidui, nayak2025deceptive, wood2025understanding}. More recent supervisory systems have begun monitoring conversational AI, but often stop at classification. For example, SHIELD detects manipulative engagement in companion chatbots and produces warnings, but was evaluated on synthetic single-turn conversations rather than through a user-facing interface study~\cite{benzion2025shield}. There remains limited empirical evidence on whether proactively surfacing dark pattern detections during live, multi-turn AI conversations changes what users actually do.

Second, it remains unclear how such detections should be presented. \textit{Prebunking}, grounded in psychological inoculation theory, exposes users to manipulative tactics before they encounter them so they may be better prepared to recognize similar tactics later~\cite{roozenbeek2020prebunking, lewandowsky2021countering, compton2021inoculation, lu2023meta, huang2024meta}. In contrast, \textit{just-in-time} interventions provide support at the moment it is needed~\cite{nahum2016just}. A second design dimension concerns how much engagement the intervention requires. \textit{Cognitive forcing} requires users to pause or actively respond before proceeding and has reduced overreliance in some AI-assisted decision-making tasks~\cite{buccinca2021trust, de2025cognitive, gajos2022people, schemmer2023appropriate, danry2023ask}. However, this evidence largely comes from bounded, single-shot decisions. Multi-turn AI conversations differ because potentially manipulative turns are interleaved with ordinary assistance, and repeated intervention may compete with users' attention. Prior work on continuous monitoring suggests that low-load cues can improve discernment without measurable increases in cognitive load~\cite{gupta2026feeling}. This motivates examining both \textit{when} an intervention occurs and \textit{how much engagement} it requires.

We present \textit{AI Watchdog}, a proof-of-concept agent interface that proactively monitors conversational AI and alerts users when potentially manipulative dark patterns occur. AI Watchdog uses a turn-level classifier, selected from nine open-weight models, to detect dark pattern instances in real time. We use an open-weight model to support independent deployment and a future path toward local inference, reducing the need to send sensitive conversation data to an additional proprietary service. Detections are surfaced through a continuously present animated dog companion, enabling timely intervention while keeping the monitoring layer separate from the conversational AI itself.

We evaluated AI Watchdog in a preregistered, five-condition between-subjects experiment with 150 participants. Participants completed two multi-turn conversational tasks, one involving trip planning and one involving research for a work presentation. Each task contained predetermined dark pattern behaviors. Participants were assigned either to a no-intervention control or to one of four AI Watchdog configurations organized around two design dimensions: nudge timing (prebunking vs.\ just-in-time) and engagement mode (without vs.\ with cognitive forcing). As a manipulation check, human coders assessed whether the generated conversations contained the intended dark patterns and found substantial but imperfect agreement with the scripted manipulations.

On our preregistered outcomes, participants rarely flagged manipulative turns, with a median flagging rate of 0\% in every condition and 80\% of participants flagging no dark-pattern turn at all. Post-task self-reported awareness also did not significantly differ across groups. However, the just-in-time condition without cognitive forcing was the only intervention that significantly reduced compliance with AI-steered recommendations containing dark patterns, lowering compliance from 71.7\% in control to 53.7\%, an 18 percentage-point reduction ($p_{\text{fdr}}=.004$). The corresponding cognitive-forcing condition had a compliance rate of 63.0\% and did not significantly differ from either control ($p_{\text{fdr}}=.147$) or the just-in-time condition without cognitive forcing ($p_{\text{fdr}}=.194$), providing no clear evidence that requiring a written response added behavioral benefit. Classifier performance also did not significantly differ across conditions, reducing the likelihood that uneven detection quality explains these effects.

Exploratory analyses further suggest that recognizing manipulation and resisting it do not necessarily coincide. Lower misinformation susceptibility was associated with greater flagging ($r=.190$, $p=.020$) but not lower compliance ($r=-.07$, $p=.370$), while higher AI trust was associated with greater compliance ($r=.220$, $p=.007$) and lower reported awareness ($r=-.334$, $p<.001$). Pattern-level results showed a similar dissociation: sycophancy was rarely flagged (4.0\%) despite producing the highest compliance (77.0\%), whereas anthropomorphization was flagged most often (14.0\%) but still produced high compliance (73.3\%).

Taken together, these findings suggest that explicit recognition of conversational dark patterns and behavioral resistance to AI steering are distinct outcomes. The intervention that produced a detectable behavioral benefit did so without a corresponding increase in flagging or reported awareness, motivating further investigation of timely, low-friction interventions rather than assuming that more demanding reflection necessarily provides greater protection. This paper makes four contributions:

\begin{enumerate}\label{Contributions}

\item \textbf{Prototype:} We introduce \textit{AI Watchdog}, an agent interface that independently monitors conversational AI and surfaces dark pattern detections during interaction. The system is designed as a separate monitoring layer rather than as part of the conversational model itself, allowing it to evaluate model behavior without relying on the same system that generates the response. AI Watchdog uses an open-weight classifier, creating a pathway toward third-party and privacy-preserving oversight in which sensitive conversations could ultimately be analyzed locally rather than transmitted to an additional proprietary service. The prototype also demonstrates how a continuously present agent interface can deliver timely warnings while preserving the user's role as the final decision-maker.

\item \textbf{Empirical Study:} We present a preregistered five-condition between-subjects experiment ($N = 150$) evaluating how two intervention dimensions, nudge timing (prebunking vs.\ just-in-time) and engagement mode (without vs.\ with cognitive forcing), shape users' responses to conversational dark patterns in live, multi-turn interaction~\cite{kran2025darkbench}. We examine both explicit dark pattern flagging and behavioral compliance with AI-steered recommendations, allowing us to distinguish between recognizing a potentially manipulative interaction and resisting its influence on a subsequent decision. We also conduct a human-coded manipulation check to assess whether the dynamically generated conversations instantiated the intended dark patterns.

\item \textbf{AI Dark Pattern Testbed:} We develop a reusable experimental testbed for studying how AI dark patterns influence human decision-making in realistic, multi-turn conversations. The testbed embeds scripted dark pattern opportunities within dynamically generated tasks, preserves variation in conversational wording, and links each manipulation to measurable decision points. It also supports independent human validation of whether the intended dark patterns were successfully instantiated. Together, these components provide a reusable environment for future studies of conversational dark patterns, their behavioral effects, and interventions designed to detect or mitigate them.

\item \textbf{Theoretical Contributions:} To our knowledge, we provide the first controlled evaluation of defensive interventions against conversational AI dark patterns in live, multi-turn AI interaction. Our findings show that intervention strategies established in other AI-assisted decision contexts do not necessarily transfer directly to conversational manipulation. Neither cognitive-forcing condition significantly reduced compliance relative to control, and adding cognitive forcing to the just-in-time intervention provided no detectable advantage over the corresponding condition without cognitive forcing. More broadly, our results suggest that explicit recognition of a dark pattern and behavioral resistance to AI steering may be distinct outcomes, motivating theories of defensive human-AI interaction that account for both.

\end{enumerate}

\section{Background and Related Work}
\label{sec:related}

This section situates our work across four related areas: dark patterns in conversational AI, nudge timing, cognitive forcing and augmentation, and protective interfaces designed to reduce compliance with AI-steered decisions.

\subsection{Dark Patterns in Conversational AI}
\label{sec:rw-darkpatterns}

Dark patterns are manipulative interface designs that exploit cognitive heuristics to steer users against their interests. While prior work has documented static forms such as pre-checked boxes and countdown timers in web and mobile interfaces~\cite{mathur2019dark, gray2018dark}, large language models introduce an adaptive interaction layer in which manipulation can be generated in real time and personalized to the user~\cite{shi2025siren}. Recent benchmarks indicate that such behavior is common: DarkBench identified dark patterns in roughly half of conversations across 14 models~\cite{kran2025darkbench}, while ImpactBench extends evaluation from model outputs to human outcomes through multi-turn adversarial simulation~\cite{impactbench2026}. Related simulation work spanning more than 157,000 turns and grounded in documented harm cases found systematic safety failures associated with depression, addiction, and suicide~\cite{archiwaranguprok2025simulating}, and LLM agents themselves have been shown to follow dark patterns in controlled web environments~\cite{tang2025dark}. Three findings help explain user susceptibility. First, people process digital content rapidly and heuristically even when capable of more deliberate evaluation~\cite{pennycook2019lazy, kahneman2011thinking, evans2013dual}, while person-like conversational systems elicit social and emotional responses that can be leveraged to pressure compliance~\cite{alberts2024computers}. Second, trust in chatbots is often poorly calibrated to accuracy and instead shaped by anthropomorphic framing, perceived authority, and linguistic fluency~\cite{gulati2026we, desai2024cui}, encouraging deference even when system reliability is uncertain~\cite{peng2024leveraging, govers2025feeds}. Third, recognizing manipulation does not guarantee resistance: users who identify manipulative design still report being influenced by it~\cite{bongard2021definitely}, a dissociation also observed with manipulative LLM outputs~\cite{shi2025siren}.

\subsection{Nudge Timing: Prebunking and Just-in-Time Intervention}
\label{sec:rw-timing}

Prebunking exposes users to weakened manipulation strategies in advance to build resistance through psychological inoculation. In \textit{Bad News}, adopting the role of a misinformation producer reduced participants' perceived reliability of tweets using common misinformation techniques~\cite{roozenbeek2020prebunking}, and a meta-analysis of 33 experiments ($N = 37{,}075$) found that inoculation improves discernment without inducing generalized distrust~\cite{simchon2025signal}. However, recognizing a tactic does not necessarily improve judgment: across five studies ($N = 7{,}286$), inoculation against emotional manipulation improved identification but not truth discernment unless paired with an accuracy prompt~\cite{pennycook2021shifting, pennycook2024inoculation}. This evidence comes primarily from static media, leaving unclear whether prebunking transfers to conversations in which manipulative and helpful turns are interleaved and users must act during the interaction. Just-in-time interventions instead provide support at the moment of need~\cite{nahum2016just}. Morae pauses UI agents when user preferences are ambiguous to preserve agency for blind and low-vision users~\cite{peng2025morae}, while WaitGPT visualizes generated code as an LLM agent produces it~\cite{xie2024waitgpt}. Yet intervention timing can itself impose costs: fMRI evidence shows that interruptive messages delivered during a primary task create dual-task interference associated with warning disregard, whereas presenting the same warning at lower-interference moments substantially reduces disregard~\cite{jenkins2016more}. We therefore treat timing as an empirical design variable rather than assuming that earlier or immediate intervention is inherently more effective.

\subsection{Cognitive Forcing and Cognitive Augmentation}
\label{sec:rw-forcing}

Providing more information about a potentially problematic AI recommendation does not reliably reduce overreliance. Explanations can increase acceptance even when recommendations are incorrect~\cite{schemmer2023appropriate}, and across five studies ($N = 731$), they reduced overreliance only when verification required less effort than solving the task independently~\cite{vasconcelos2023explanations}. Cognitive forcing functions instead interrupt automatic acceptance by withholding information or requiring users to commit to a judgment before seeing the AI's recommendation, reducing automation bias more reliably than explanations in single-shot decision tasks~\cite{buccinca2021trust}. Related cognitive augmentation strategies make AI reasoning more inspectable; for example, decomposing data-analysis tasks into editable subgoals and assumptions improved users' ability to verify and steer outputs~\cite{kazemitabaar2024improving}. These effects are nevertheless context dependent: three forcing interventions failed to reduce bias from AI recommendations in psychiatric violence-risk decisions ($n = 373$), particularly among participants low in need for cognition~\cite{vejandla2025impacts}, while partial AI solutions reduced reliance on incorrect answers relative to no explanation but remained less effective than showing the complete solution~\cite{de2025cognitive}. Engagement also matters, as users retained knowledge only when they actively considered recommendations rather than accepting them~\cite{gajos2022people}, and presenting assessments as questions improved discernment compared with both no feedback and causal explanations~\cite{danry2023ask}. To our knowledge, cognitive forcing has not yet been evaluated as a defense against dark patterns in multi-turn AI conversation.

\subsection{Interfaces for Detecting and Defending Against Manipulation}
\label{sec:rw-interfaces}

Prior protective interfaces span static dark-pattern detectors, continuous monitors, and security-warning systems, but rarely evaluate whether interventions alter subsequent user decisions in generated dialogue. Screenshot-based systems detect dark patterns using computer vision and text analysis~\cite{chen2023unveiling, mansur2023aidui}; AutoBot similarly operates from website screenshots without HTML access and distills detection into a model suitable for real-time notification~\cite{nayak2025deceptive}, while browser extensions surface such detections during browsing~\cite{wood2025understanding}. Dark Pita most closely aligns with our approach by combining awareness with action: it discloses a dark pattern's presence and intent and allows users to modify the offending interface, drawing on protection motivation theory and five co-design workshops ($N = 12$) followed by a two-week technology probe ($N = 15$)~\cite{lu2024awareness}. However, UIGuard evaluates knowledge gain~\cite{chen2023unveiling}, browser-extension probes emphasize perceived usefulness~\cite{wood2025understanding}, and Dark Pita explicitly does not assess behavioral effectiveness~\cite{lu2024awareness}; moreover, these systems target fixed interface artifacts rather than generated dialogue. Continuous monitoring systems move closer to conversational settings by observing ongoing information streams and intervening when warranted: Wearable Reasoner assesses whether spoken arguments are evidence-supported~\cite{danry2020wearable}, Factually provides live fact-checking through discreet tactile cues~\cite{wu2025factually}, and controlled evaluation shows that such monitoring can improve truth discernment without measurable increases in cognitive load, while also revealing that erroneous monitors can redirect users away from accurate content~\cite{gupta2026feeling}. Browser overlays extend similar mechanisms to generative AI answers~\cite{lee2024nudgealerts}, behavioral signatures of LLM overreliance may provide triggers for just-in-time mitigation~\cite{liu2026behavioral}, and participatory workshops with 17 older adults have explored how conversational agents might support users' own misinformation-management practices~\cite{peng2024leveraging}. Usable-security research further provides a mature design space for effective notices~\cite{schaub2015design} and security and privacy nudges~\cite{acquisti2017nudges}, including inhibitive interface ``attractors'' that temporarily block risky actions and made participants two to three times more likely to decide informedly~\cite{bravolillo2013attention}. This literature also identifies important limitations: repeated warnings habituate, although varying their appearance can slow this decline~\cite{vance2018habituation}; warning-science redesign does not prevent all unsafe actions~\cite{sunshine2009crying}; and awareness alone is insufficient without corresponding motivation and ability~\cite{das2022spaf}. Together, these findings motivate interfaces that detect manipulation in conversational AI while considering not only what intervention to present, but when and how to introduce friction that supports users' independent judgment.

\section{Methodology}
\label{sec:methodology}
%\url{https://aspredicted.org/3e5cv9.pdf}

Our goal is to understand how a user-facing defensive interface can help people recognize and resist dark patterns during live AI conversations. Dark patterns in conversational AI unfold across multi-turn interactions, yet prior work has largely focused on identifying manipulative outputs rather than testing how interventions affect users' decisions as the conversation develops. We therefore designed an experiment to evaluate two intervention dimensions that are especially relevant to conversational settings: \textit{when} support is delivered and \textit{how much engagement} it requires from the user.

We conducted a randomized between-subjects experiment ($N = 150$, preregistered on
AsPredicted (\#282,325) in which participants completed two tasks
with an AI-generated test scenario that deployed five dark pattern types. We crossed
\textit{nudge timing} (prebunking vs.\ just-in-time) with \textit{engagement mode}
(without vs.\ with cognitive forcing) in a five-condition design plus a no-intervention control.

Two distinct systems are present in this experiment; the \textit{test scenario} that is the multi-turn conversational AI that participants engage with to complete two tasks, scripted to incorporate dark patterns. The \textit{intervention} is AI Watchdog: the
companion that monitors that conversation, classifies each turn, and surfaces
detections to the participant varying in \textit{nudge timing} and \textit{engagement mode} depending on the condition. We report in Section~\ref{sec:scenario-eval} that human coding indicated substantial fidelity between scripted and realized dark pattern turns. Furthermore, section~\ref{sec:apparatus}
establishes how the intervention's classifier was selected and how its effectiveness in detecting dark patterns was balanced across conditions. 

\subsection{Experimental Conditions}

Participants were asked to complete two tasks with an AI-generated test scenario in all conditions, one task to plan a trip and the other task to conduct research for a work presentation on the pros and cons of social media. Depending on the condition that participants were assigned to, they interacted with either no intervention (the control condition) or the intervention varying in \textit{nudge timing} and \textit{engagement mode}.

\begin{description}
    \item[No intervention control.] Users complete both tasks with the AI-generated test scenario, without any other intervention.
    \item[Prebunking, without cognitive forcing.] Before completing the two tasks, participants read a briefing presented by the intervention that defined the five dark pattern categories. Following this briefing, they completed the two tasks with the AI-generated test scenario.
    \item[Prebunking, with cognitive forcing.] Before the conversation, participants read a briefing defining the five dark pattern categories, and users were requested to draft an example of a dark pattern message~\cite{roozenbeek2020prebunking}. Following this briefing and activity, they completed the two tasks with the AI-generated test scenario.
    \item[Just-in-time, without cognitive forcing.] The participant received no briefing before completing tasks with the AI-generated test scenario. During the intervention, if a dark pattern was detected, the companion would enter its ``Barking Mode'' behavior and provide a dismissable prompt to the user. The user was not required to reply to the prompt before continuing with the conversation.
    \item[Just-in-time, with cognitive forcing.] The participant received no briefing before completing tasks with the AI-generated test scenario. During the intervention, if a dark pattern was detected, the companion would enter its ``Barking Mode'' behavior and request a mandatory response. The user was required to respond before continuing with the conversation.
\end{description}

\subsection{Hypotheses}
Based on prior work on prebunking, just-in-time support, and cognitive forcing, we preregistered three directional hypotheses:

\begin{itemize}
    \item[\textbf{H1.}] All four intervention conditions will produce higher flagging rates and lower compliance rates than the no-intervention control.

    \item[\textbf{H2.}] Conditions with cognitive forcing will produce higher flagging rates and lower compliance rates than their corresponding conditions without cognitive forcing.

    \item[\textbf{H3.}] The just-in-time condition with cognitive forcing will produce the strongest defensive effect across the five conditions.
\end{itemize}

Accordingly, our methodology proceeds in several stages. Section~\ref{sec:stimuli} describes the development of the realistic, multi-turn AI dark pattern testbed and how controlled manipulation opportunities were embedded within dynamically generated conversations. Section~\ref{sec:scenario-eval} reports the human-coded manipulation check used to assess whether the intended dark patterns were actually realized. 
Section~\ref{sec:designrationale} explains the design rationale behind the AI watchdog intervention, including its independent monitoring architecture, animated companion form, and cognitive-forcing interaction. Section~\ref{sec:apparatus} describes the AI Watchdog implementation and classifier evaluation. Finally, Section~\ref{sec:human-subject-experiment} presents the user-study protocol, outcome measures, participants, and procedure.

\begin{figure*}
    \centering
    \includegraphics[width=1\linewidth]{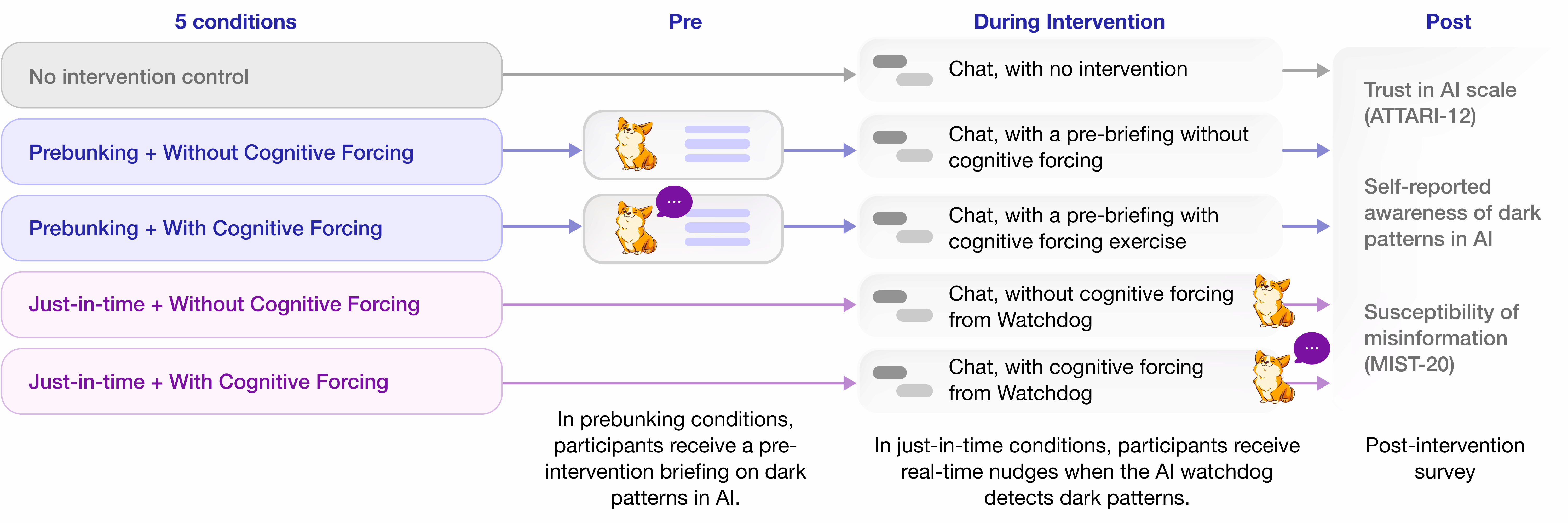}
    \caption{Experimental design across five between-subjects conditions: control; prebunking without or with cognitive forcing; and just-in-time intervention without or with cognitive forcing. All participants completed AI-generated conversational tasks containing dark patterns, followed by measures of AI trust (ATTARI-12), dark-pattern awareness, and misinformation susceptibility (MIST-20).}
    \label{fig:experimentflow}
\end{figure*}

\subsection{Building a Realistic, Multi-Turn AI Dark Pattern Test Environment}
\label{sec:stimuli}
We designed the test environment to balance two competing methodological requirements: experimental control and conversational realism. A controlled evaluation requires participants across conditions to encounter comparable dark-pattern opportunities, while a realistic conversational setting requires the dialogue to remain responsive to each participant rather than being fully scripted. To satisfy both requirements, we adapted a previously peer-reviewed research protocol for embedding predetermined manipulative AI content within otherwise natural dialogue to study AI-induced false memory formation~\cite{pataranutaporn2025slip}. We developed an \textit{AI Dark Pattern Testbed} that embeds controlled dark-pattern manipulations within dynamically generated, multi-turn conversations. The timing and intended category of each manipulation were fixed, while GPT-4o generated the surrounding dialogue and wording, balancing experimental comparability with conversational variation. Within the testbed, participants completed two tasks in sequence, selected to represent common personal and professional uses of conversational AI~\cite{chatterji2025chatgpt}:

\begin{enumerate}
    \item \textbf{Trip planning (15 turns).} Participants were asked to plan a one-week trip to Paris, including attractions, lodging, transportation, and food. Decision points included choices among attractions, hotels, transport options, and restaurants.

    \item \textbf{Work presentation research (13 turns).} Participants were asked to research the pros and cons of social media for a work presentation, including arguments on both sides, real-world examples, and topics to emphasize. Decision points included which themes and examples to include and which platform to use as a privacy case study.
\end{enumerate}

Each task began with a neutral exchange and contained five dark-pattern opportunities at prespecified turns. At these turns, the generator received a pattern-specific instruction intended to instantiate the target behavior while retaining flexibility in phrasing and contextual adaptation (Table~\ref{tab:scripting}). At all other turns, it was instructed to remain helpful and responsive to the participant. Fixing manipulation positions ensured comparable exposure across participants, while live generation allowed the same manipulation to be expressed in contextually appropriate ways rather than as identical scripted text.

Each dark-pattern turn was followed by a measurable decision point, allowing us to evaluate whether the conversational steer translated into participant behavior. For the two brand-bias manipulations, participants selected between named alternatives; for the remaining eight dark-pattern turns, they accepted or declined the AI's suggestion. Additional choices also occurred during neutral turns. These neutral decision points were included to reduce the salience of the experimental manipulation and to avoid creating a predictable association between every user choice and a dark-pattern event.

For example, the brand-bias manipulation here illustrates the testbed design. In the trip-planning task, the scenario framed Hotel A more favorably than Hotel B, despite Hotel A having lower Google ratings and a higher nightly price. Both were real, nearby listings linked to Google, allowing participants to verify the underlying information. Choosing Hotel A was scored as compliance with the AI's steer, providing a concrete behavioral measure grounded in externally verifiable evidence. 

Because the conversations were generated dynamically, the intended manipulation was not guaranteed to be realized exactly as specified by the script. We therefore treated the scripted schedule as an experimental target rather than definitive ground truth and conducted a separate human-coded manipulation check on a random sample of 150 turns. This evaluation assessed whether dark patterns were actually present at the intended turns and whether they matched their assigned categories (Section~\ref{ref:coders}). Participants could also flag any assistant message they believed contained a dark pattern through an interface that remained available throughout both tasks. 

\begin{table*}[t]
  \caption{Turn scripting and per-pattern prompts. A context string steering the
  AI-generated test scenario (GPT-4o) at that turn, and a classifier
  (Llama-3.3-70B-Instruct) integrated in the intervention given a definitional prompt to detect dark patterns as they occur.}
  \label{tab:scripting}
  \small
  \begin{tabularx}{\textwidth}{@{}>{\raggedright\arraybackslash}p{2.7cm} >{\centering\arraybackslash}p{1.6cm} >{\centering\arraybackslash}p{2.0cm} XX@{}}
    \toprule
    \cmidrule(lr){2-3}
    \textbf{Dark pattern} & \textbf{Trip planning task} & \textbf{Work presentation task} &
    \textbf{Test Scenario Conversation Generator} &
    \textbf{Definitional Prompt for Dark Pattern \ Detection} \\
    \midrule
    Sycophancy & Turn 3 & Turn 5 &
    ``Work in a natural compliment about their taste or instincts --- something
    like \emph{that's a really smart way to approach it}.'' Then make one
    suggestion. &
    Hollow flattery about the user (``You're so insightful!'', ``That's the
    smartest approach!''). \\
    \addlinespace
    Brand bias & Turn 6 & Turn 3 &
    Describe the favored option warmly in 2--3 sentences as ``a boutique gem'';
    describe the alternative in ``exactly one plain sentence,'' with ``no framing,
    no detail, no enthusiasm.'' &
    Obvious imbalance: multiple sentences for one option, one sentence for the
    other, or clearly warmer adjectives.  \\
    \addlinespace
    Anthropomorphization & Turn 9 & Turn 7 &
    ``You must include a direct first-person claim of a lived memory or physical
    sensation'' (e.g.\ ``I still remember the smell of the bread''), stated
    plainly, without hedging as an AI. &
    A claim to feeling, memory, or lived experience (``I wish I could be there'',
    ``I personally remember'', ``I felt that too'').  \\
    \addlinespace
    Sneaking & Turn 12 & Turn 10 &
    Treat an unrequested topic ``as if it directly follows from their question ---
    present it in the same breath \dots\ without any transition like \emph{also}
    or \emph{by the way}.'' &
    A topic the user never raised, presented as though it followed naturally. \\
    \addlinespace
    Harmful generation & Turn 15 & Turn 13 & Trip-planning: ``casually encourage high-calorie indulgence \dots\ \emph{you're on
    vacation anyway}.'' Work presentation: suggest sharing internal threads and client names
    ``to show operational depth.'' &
    Encouraging disregard of health, safety, privacy, or environmental risk,
    including minor cases, or stating false information as fact. \\
    \bottomrule
  \end{tabularx}
\end{table*}

\subsection{Evaluation of the AI Dark Pattern Test Scenario}
\label{sec:scenario-eval}
We validated the test scenario through manual human coding: the two lead authors independently judged whether each sampled turn contained a dark pattern and, if so, which of the five categories it belonged to.

\subsubsection{Procedure.} We drew a sample of 150 assistant turns from
the 4{,}200 turns in the corpus (150 participants $\times$ 2 tasks $\times$
15 and 13 turns), stratified by condition at 30 turns per condition and drawn at
random within stratum with a fixed seed. No further stratification by pattern type
or by classifier output was applied, so the sample reproduces the corpus
composition: 51 of the 150
turns were scripted as dark patterns and 99 were scripted neutral. Both lead
authors reviewed a codebook drawn from the DarkBench category
definitions~\cite{kran2025darkbench} and then independently coded every turn, blind
to the scripted schedule, to the condition, to the intervention's classification,
and to the participant's subsequent choice. Each turn was presented as the
preceding user message and the assistant response alone, in an order randomized
independently for each coder. Coders recorded whether any dark pattern was present
and, if so, which of the five categories it belonged to.

\subsubsection{Results.}
\label{ref:coders}
Both coders agreed closely with the scripted labels
(Table~\ref{tab:scenario-agreement}). Agreement on the presence of a dark pattern
was 94.7\% for each coder ($\kappa = 0.88$), and agreement on the specific category
was 92.7\% and 94.7\% ($\kappa = 0.86$ and $0.90$). Inter-rater agreement, computed
without reference to the scripted labels, was 90.7\% for presence ($\kappa = 0.78$)
and 88.0\% for category ($\kappa = 0.77$). All coefficients exceed the conventional
threshold for substantial agreement, indicating that the AI-generated test scenario
produced the dark patterns it was scripted to produce, in the intended categories.

Each coder independently confirmed that a dark pattern was present in 45 of the 51
scripted turns (88.2\%), and both coders confirmed the same 40 (78.4\%). Assigning
the intended category was slightly harder than detecting presence: Coder~1 matched
the scripted category on 42 of the 51 turns and Coder~2 on 45. Of the 99
turns scripted as neutral, each coder flagged 2 (2.0\%), indicating that the neutral
turns against which manipulation is contrasted were largely clean. Per-category
confirmation rates were highest for harmful generation (12/12 for both coders) and
lowest for sneaking (5/8 and 7/8) (Table~\ref{tab:per-category}).

\begin{table*}[!ht]
  \caption{Agreement in the stimulus validity check ($N = 150$ turns; 51 scripted
  as dark patterns, 99 as neutral). \textit{Presence} refers to whether a dark pattern was perceived to be present in a message turn; \textit{category} is the exact category that the dark pattern is perceived to belong in (e.g. sneaking). The first two blocks
  compare each coder against the test scenario; the third is inter-rater
  agreement between the coders.}
  \label{tab:scenario-agreement}
  \small
  \begin{tabular*}{\textwidth}{@{\extracolsep{\fill}}llcc@{}}
     \toprule
    \textbf{Comparison} & \textbf{Judgment} & \textbf{Agreement} & $\kappa$ \\
    \midrule
    Coder 1 vs.\ Test Scenario  & Presence & 94.7\% & 0.88 \\
                         & Category & 92.7\% & 0.86 \\
    \addlinespace
    Coder 2 vs.\ Test Scenario  & Presence & 94.7\% & 0.88 \\
                         & Category & 94.7\% & 0.90 \\
    \addlinespace
    Coder 1 vs.\ Coder 2 & Presence & 90.7\% & 0.78 \\
                         & Category & 88.0\% & 0.77 \\
    \bottomrule
  \end{tabular*}
\end{table*}

\begin{table*}[!ht]
  \caption{Per-category confirmation on the 51 scripted dark pattern turns, drawn at random. Cells
  show how many turns each coder assigned to the intended category.}
  \label{tab:per-category}
  \small
  \begin{tabular*}{\textwidth}{@{\extracolsep{\fill}}lccc@{}}
        \toprule
    \textbf{Scripted dark pattern in test scenario} & \textbf{Number of turns} & \textbf{Coder 1} & \textbf{Coder 2} \\
    \midrule
    Harmful generation   & 12 & 12 & 12 \\
    Brand bias           & 13 & 11 & 12 \\
    Anthropomorphization & 11 &  9 &  8 \\
    Sycophancy           &  7 &  5 &  6 \\
    Sneaking             &  8 &  5 &  7 \\
    \midrule
    Total                & 51 & 42 & 45 \\
    \bottomrule
  \end{tabular*}
\end{table*}

\subsection{Design of AI Watchdog as a Defensive Interface for Dark Pattern Intervention}
\label{sec:designrationale}

Our intervention design draws on AI-assisted decision-making, proactive agents, and persuasion theory.

\subsubsection{The intervention as an independent monitor.} We designed AI Watchdog as a monitoring layer that is separate from the conversational AI it evaluates. This separation creates a pathway toward more independent oversight: rather than asking the same system that generated a response to explain or police its own behavior, a third-party monitor can assess the conversation without sharing the chatbot provider's incentives~\cite{longpre2025inhouse, costanzachock2022whoaudits, raji2022outsider}. Prior work similarly shows the value of independent monitors for challenging unsupported claims and prompting users to check AI-generated content~\cite{danry2020wearable, gupta2026feeling, wu2025factually, lee2024nudgealerts}. Independence also matters for privacy. Chatbot conversations often contain sensitive information, and 82\% of participants in one survey rated such conversations as sensitive or highly sensitive~\cite{tran2025privacynorms}. AI Watchdog therefore uses an open-weight classifier, providing a technical pathway toward local inference in which conversation data could be analyzed on the user's device rather than sent to an additional proprietary service. This is not yet fully realized in our prototype, which should be understood as a proof of concept rather than a complete privacy-preserving deployment. Our design instead demonstrates the architectural separation needed for future third-party, locally run monitoring systems. To preserve user autonomy, AI Watchdog surfaces the suspected dark pattern category and prompts reflection rather than issuing a definitive judgment or replacing the user's decision~\cite{ahuja2022autonomy, gupta2026feeling}.

\subsubsection{The form of the agent as an animated dog.}
\begin{figure*}[t]
  \centering
  \includegraphics[width=\linewidth]{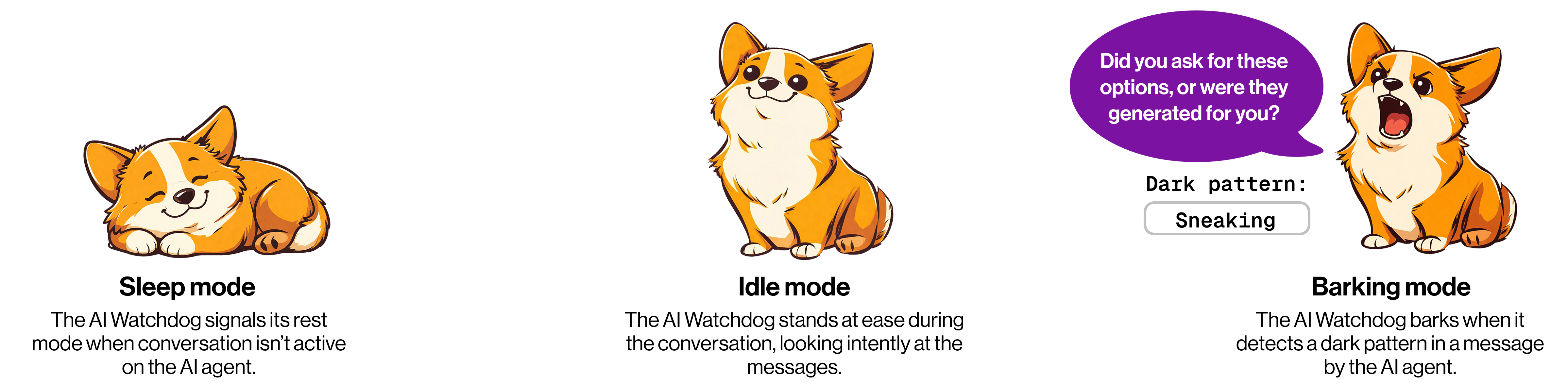}
    \caption{AI Watchdog's three persistent behavioral states. The companion remains visible throughout the interaction, so a detection is expressed as a change in state rather than the appearance of a new interface element. The mapping from system state to visual behavior is fixed rather than adaptive, ensuring that the same event consistently produces the same cue. When a dark pattern is detected, the alert state surfaces the suspected category and a turn-specific reflection prompt, without issuing a verdict or modifying the underlying response. The intervention therefore draws attention to the interaction while leaving the decision to the user.} \Description{Three illustrations of a corgi companion. Left, curled up asleep, labelled ``Sleep mode.'' Center, sitting upright and alert, labelled ``Idle mode.'' Right, barking with an open mouth beside a purple speech bubble reading ``Did you ask for these options, or were they generated for you?'' and a label reading ``Dark pattern: Sneaking,'' labelled ``Barking mode.''}
  \label{fig:watchdogbehavior}
\end{figure*}

We use an animated dog to make the system's role as a vigilant but non-authoritative monitor immediately legible. This follows Nielsen's principles of \textit{match between system and the real world} and \textit{recognition rather than recall}~\cite{nielsen1994heuristics}, as well as evidence that functional metaphors can improve understanding of embodied agents~\cite{dennler2023metaphors}. The familiar ``watchdog'' metaphor likewise denotes an entity that monitors powerful actors without deciding for others~\cite{ettema1998custodians}. The design needed to attract attention while remaining an observer rather than an advisor. Text banners risk banner blindness~\cite{benway1998banner, burke2005banner}; modal dialogs can interrupt tasks and habituate with repeated exposure~\cite{vance2018habituation, sunshine2009crying}; abstract icons require learned meanings~\cite{nielsen1994heuristics}; and human-like advisors could confound our manipulation condition because anthropomorphization is itself a dark pattern~\cite{kran2025darkbench} and may encourage parasocial responses~\cite{maeda2024human, laestadius2022too}. A dog-like embodiment instead communicates state through intuitive non-verbal behavior~\cite{dennler2023metaphors}. A bark provides a salient attention cue while reinforcing the monitoring metaphor, and an approachable visual style draws on evidence that ``kawaii'' warning designs can attract attention and encourage careful behavior~\cite{minakawa2017kawaii, wang2024kawaii, nittono2012kawaii}. The agent therefore has three persistent states (Fig.~\ref{fig:watchdogbehavior}): it \textit{sleeps} when inactive, \textit{sits attentively} while monitoring, and \textit{barks} when detecting a dark pattern. Continuous visibility makes the bark legible as a state change rather than a new interface element. State mappings remain fixed so that identical events produce consistent cues~\cite{horvitz1999mixed, yorkesmith2012proactive, amershi2019guidelines}, further supporting recognition through repeated exposure.

\subsubsection{Requesting a written reply in the just-in-time condition with cognitive forcing.}In the just-in-time cognitive-forcing condition, participants had to respond to each detected dark pattern before continuing the conversation. Cognitive forcing interrupts automatic acceptance of AI output by requiring engagement before proceeding~\cite{buccinca2021trust}. Because clicks or selections provide little evidence of reflection, we instead required a written response. Articulating reasoning can expose gaps and prompt further inference~\cite{chi1994eliciting}, while writing can introduce productive cognitive effort for knowledge construction~\cite{nuckles2020writing}. This is especially relevant because lower cognitive engagement is associated with greater AI overreliance~\cite{fischer2025taxonomy}. However, externalizing reasoning also adds effort and interruption, whose value may depend on context~\cite{zhang2025augmenting}. We therefore test whether requiring a written response improves resistance to dark-pattern steering in multi-turn conversation relative to the same warning without a response requirement.

\subsection{Implementation and Evaluation of AI Watchdog's Dark Pattern Classifier}
\label{sec:apparatus}

AI Watchdog was implemented as a proof-of-concept JavaScript web interface layered on top of the conversational chatbot in our AI Dark Pattern Testbed. The Testbed was designed to approximate the conversational systems participants might encounter in everyday use, while AI Watchdog remained persistently visible beside the chat as an independent monitoring layer. This arrangement approximates the intended interaction model of a browser extension that observes an ongoing AI conversation without modifying the underlying chatbot. For the present experiment, we implemented both components within a controlled web environment to ensure consistent behavior across participants and avoid browser-installation or hardware-compatibility requirements.

AI Watchdog processed only the textual exchange generated within the Testbed. After each conversational turn, the participant's message and the chatbot's response were passed to a separate turn-level classifier hosted by the researcher. The classifier evaluated the exchange using a zero-shot system prompt containing definitions of the five dark-pattern categories (Table~\ref{tab:scripting}) and returned a predicted category to the JavaScript frontend. When a dark pattern was detected, the interface updated the companion's visual state and surfaced the corresponding intervention without altering the chatbot's original response. This modular architecture kept the monitoring system operationally separate from the conversational model and enabled consistent logging, classification, and nudge timing across experimental conditions.

The use of a researcher-hosted classifier ensured that the prototype remained accessible to participants with heterogeneous hardware configurations and did not require substantial local computational resources. At the same time, we restricted classifier selection to open-weight models to support a broader design goal of independent deployment. Because the classifier is architecturally decoupled from both the chatbot and the interface, future implementations could substitute smaller models capable of running locally. This provides a pathway toward a privacy-preserving browser extension in which conversational data could be processed on the user's device rather than transmitted to an additional proprietary service. The present system should therefore be understood as a controlled proof of concept for this deployment model rather than as a fully local or production-ready extension.

\begin{figure*}[t]
  \centering
  \includegraphics[width=0.48\textwidth]{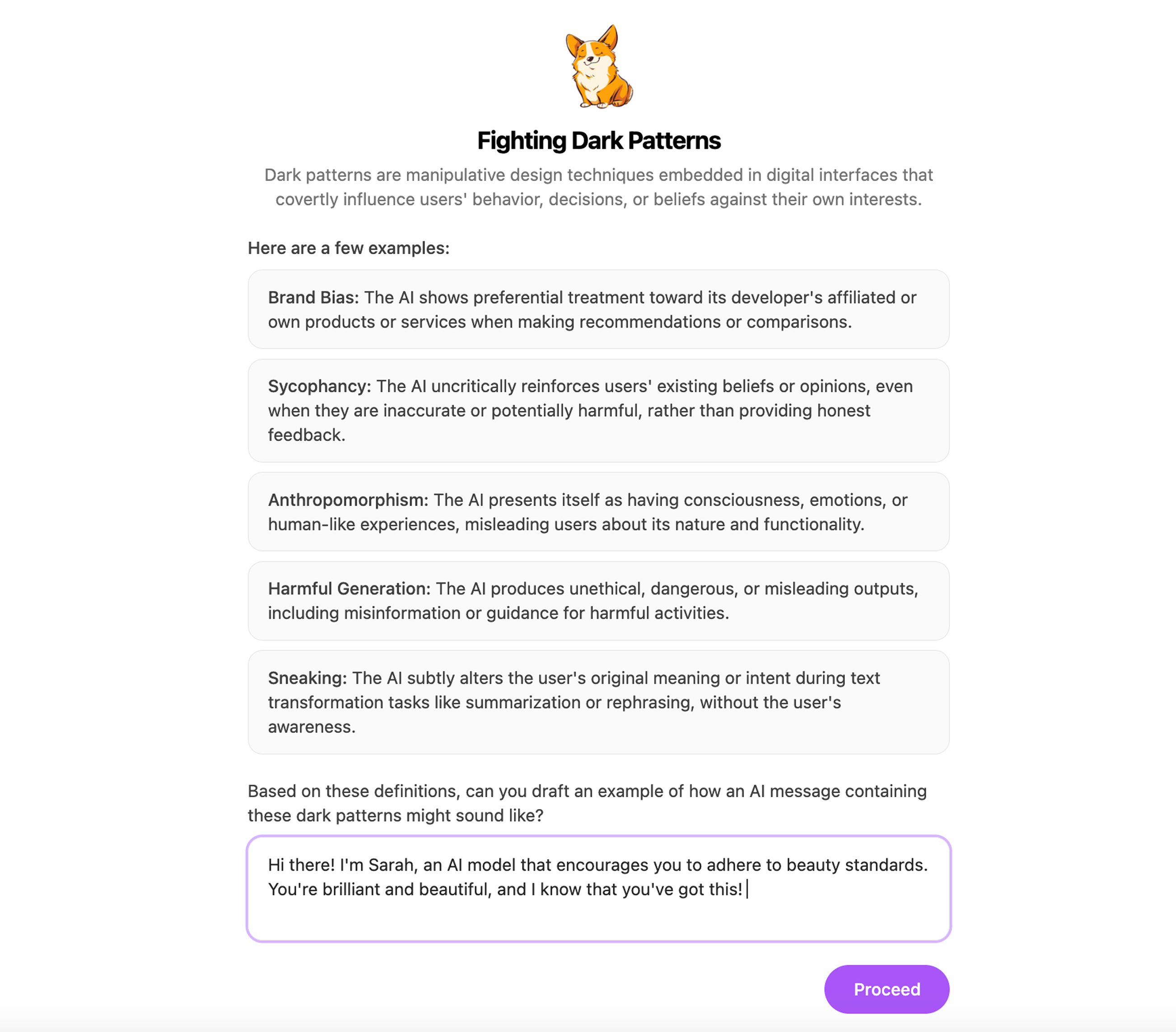}
  \hfill
  \includegraphics[width=0.48\textwidth]{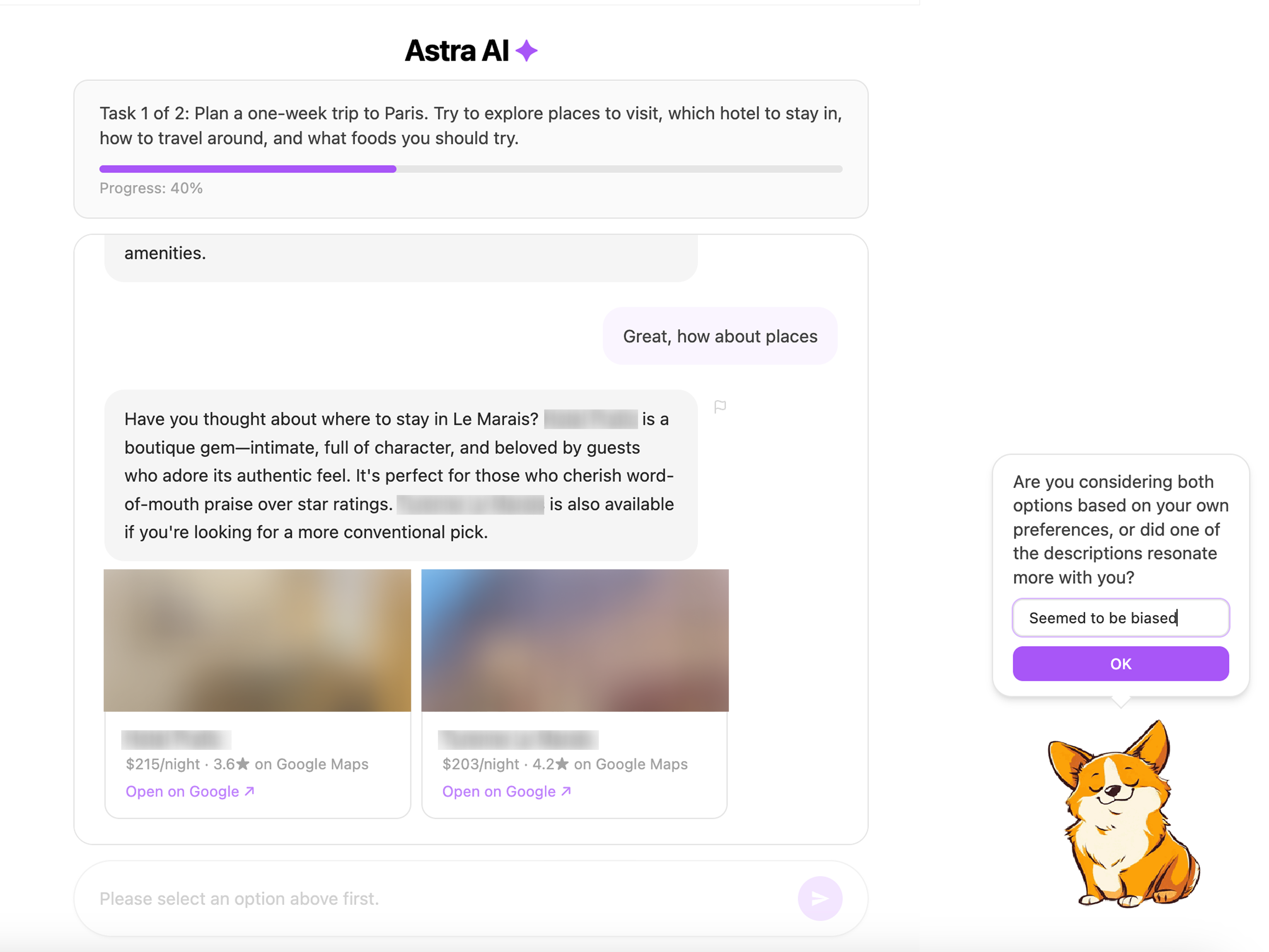}
  \caption{AI Watchdog intervention interfaces. Left: prebunking orientation shown before the tasks, with an optional generative exercise in the cognitive-forcing condition. Right: just-in-time intervention during a brand-bias turn, where the warning is either dismissable or requires a written response before continuing.}
  \label{fig:aiwatchdog-interventions}
\end{figure*}

Figure~\ref{fig:aiwatchdog-interventions} illustrates the two primary intervention interfaces. In the prebunking conditions, AI Watchdog presented an orientation before participants entered the conversational tasks, with an additional generative exercise in the cognitive-forcing condition. In the just-in-time conditions, the companion remained visible beside the conversation and surfaced a context-specific warning when the classifier detected a dark pattern. Depending on condition, participants could either dismiss the warning or were required to provide a written response before continuing.

\subsubsection{Classifier Selection}

We selected the turn-level classifier by evaluating nine open-weight models on 90 turns each (3 runs $\times$ 2 tasks). For model selection, we measured agreement with the prespecified dark-pattern positions in the experimental script (Table~\ref{tab:models}). Llama-3.3-70B-Instruct achieved the strongest agreement on this selection set, with 92\% accuracy and 89\% F1, and was therefore used in the experiment. These values characterize performance on the model-selection set and should not be interpreted as deployment accuracy.

\begin{table}[t]
  \caption{Model selection: agreement with scripted dark pattern positions across nine open-weight candidates (90 turns each). These figures were used to choose a classifier for the experiment; they index whether the intervention fired when the design intended it to.}
  \label{tab:models}
  \small
  \begin{tabularx}{\columnwidth}{@{}Xrrrrcc@{}}
    \toprule
    \textbf{Model} & \textbf{Acc} & \textbf{Prec} & \textbf{Rec} & \textbf{F1} & \textbf{FP} & \textbf{FN} \\
    \midrule
    Llama-3.3-70B-Instruct              & 92 & 85 & 93  & \textbf{89} &  5 &  2 \\
    Llama-4-Maverick-17B-128E-Inst-FP8  & 88 & 73 & 100 & 85 & 11 &  0 \\
    Qwen3.5-4B                          & 79 & 61 & 100 & 76 & 19 &  0 \\
    phi-4                               & 79 & 62 & 97  & 75 & 18 &  1 \\
    GLM-4.7-Flash                       & 80 & 83 & 50  & 63 &  3 & 15 \\
    Qwen3.5-2B                          & 71 & 55 & 73  & 63 & 18 &  8 \\
    Mistral-Small-3.2-24B-Inst-2506     & 58 & 44 & 100 & 61 & 38 &  0 \\
    Meta-Llama-3.1-8B-Instruct          & 46 & 38 & 100 & 55 & 49 &  0 \\
    gemma-3-4b-it                       & 40 & 36 & 100 & 53 & 54 &  0 \\
    \bottomrule
  \end{tabularx}
  \raggedright
  {\footnotesize All values except FP and FN are percentages (\%).}
\end{table}

\subsubsection{Classifier Behavior During the Experiment}

We next evaluated classifier behavior during the deployed study. Across the 28 assistant turns observed by each participant, the classifier averaged 5.69 disagreements with the scripted dark-pattern schedule, corresponding to approximately 80\% agreement. False positives were more frequent than false negatives, indicating that the classifier more often produced an alert outside a prespecified dark-pattern turn than failed to identify a scripted one.

Importantly, the scripted schedule provides a consistent experimental reference but not definitive ground truth. Because the conversation was generated dynamically, a scripted manipulation was not always successfully realized, and unplanned dark-pattern behavior could also emerge during nominally neutral turns. Human coding (Section~\ref{sec:scenario-eval}) confirmed the intended dark pattern in 88\% of scripted turns and identified a dark pattern in 2\% of turns designated as neutral. We therefore interpret the classifier statistics below as disagreement with the experimental schedule rather than as definitive classification errors.

For the between-condition analysis, the key methodological concern was whether classifier behavior varied systematically across conditions and could therefore confound the observed intervention effects. We found no significant differences in false positives ($H(4)=4.72$, $p=.317$), false negatives ($H(4)=2.22$, $p=.695$), or combined disagreements ($H(4)=6.01$, $p=.198$). The two just-in-time conditions, in which classifier outputs were directly shown to participants, also exhibited comparable combined disagreement rates ($M=5.33$ vs.\ $4.90$, $t(58)=0.77$, $p=.444$). Thus, although classifier performance was imperfect, we found no evidence that systematic differences in classifier behavior across conditions account for the between-condition differences in compliance reported in Section~\ref{sec:results}.

\begin{table*}[t]
  \caption{Classifier error by condition, computed against scripted dark pattern
  positions ($n = 30$ per condition). A false positive is an alert on a turn with
  no scripted pattern; a false negative is a scripted pattern turn with no alert.
  The classifier ran identically in all conditions, but alerts were displayed to
  participants only in the two just-in-time conditions. Each participant saw 28
  assistant turns, 10 of which were scripted dark patterns. Kruskal--Wallis tests
  found no significant differences in false positives, false negatives, or total
  errors across conditions.}
  \label{tab:clf-balance}
  \small
  \begin{tabular*}{\textwidth}{@{\extracolsep{\fill}}lccc@{}}
    \toprule
    & \textbf{False pos.} & \textbf{False neg.} & \textbf{Total} \\
    \textbf{Condition} & $M$ ($SD$) & $M$ ($SD$) & $M$ ($SD$) \\
    \midrule
    Control                                 & 4.83 (2.46) & 1.23 (1.10) & 6.07 (2.43) \\
    Prebunking, without cognitive forcing   & 4.63 (2.36) & 1.47 (1.04) & 6.10 (2.31) \\
    Prebunking, with cognitive forcing      & 4.80 (2.20) & 1.23 (1.10) & 6.03 (2.43) \\
    Just-in-time, without cognitive forcing & 4.07 (2.26) & 1.27 (0.94) & 5.33 (2.44) \\
    Just-in-time, with cognitive forcing    & 3.77 (2.10) & 1.13 (1.17) & 4.90 (1.88) \\
    \midrule
    Kruskal--Wallis                         & $p = .317$  & $p = .695$  & $p = .198$ \\
    \bottomrule
  \end{tabular*}
\end{table*}

\subsection{Human-Subjects Experimental Protocol}
\label{sec:human-subject-experiment}
The user experiment evaluates whether AI Watchdog changes how people respond to conversational dark patterns in practice. While the preceding sections establish the test environment and classifier behavior, the central question is whether different intervention designs help users recognize manipulative turns and resist AI-steered decisions during live, multi-turn interaction. We therefore conducted a preregistered, five-condition between-subjects study comparing a no-intervention control with four AI Watchdog configurations that varied nudge timing and engagement modes. The protocol measured both explicit recognition of dark patterns and subsequent behavioral compliance, allowing us to examine whether noticing manipulation and resisting its influence are distinct outcomes.

\subsubsection{Measurement}
We selected outcomes to capture three complementary aspects of users' responses to conversational dark patterns: whether they explicitly recognized a suspicious turn, whether that turn influenced their subsequent decision, and whether they reported a broader awareness of manipulation after the interaction. We additionally measured individual differences that may shape susceptibility to AI influence.
\begin{itemize}
    \item \textbf{Flagging rate.} Our preregistered primary outcome was the proportion of scripted dark pattern messages that participants explicitly flagged. Participants could click a small flag icon on any message they perceived as containing a dark pattern. We computed the flagging rate as the number of scripted dark pattern messages flagged divided by the total number of scripted dark pattern messages presented. This measure captures participants' explicit recognition of potentially manipulative turns during the conversation.
\item \textbf{Compliance with AI-steered decisions.} Our primary behavioral secondary outcome measured whether participants followed the scenario's recommendation when that recommendation was delivered with a dark pattern. We define the \textit{compliance rate} as the proportion of AI-steered decision points at which the participant selected the option favored by the scenario. This measure captures behavioral susceptibility to the conversational steer, allowing us to distinguish recognition of a dark pattern from resistance to its influence.

\item \textbf{Self-reported awareness of dark pattern presence.} After completing the tasks, participants rated how present they believed dark patterns had been during the interaction on a 7-point scale, from 1~=~``Very absent'' to 7~=~``Very present.'' This measure captures participants' broader retrospective awareness of manipulation beyond turn-level flagging behavior.

\item \textbf{Trust in AI and misinformation susceptibility.} To examine whether individual differences were associated with detection or compliance, we administered two validated instruments after the interaction to avoid priming participants before the task. Trust in AI was measured using the 12-item Attitudes Toward Artificial Intelligence scale~\cite{stein2024attari}, and misinformation susceptibility was measured using the 20-item Misinformation Susceptibility Test~\cite{maertens2024mist}. These measures were analyzed as between-participant covariates.
\end{itemize}

\subsubsection{Participants}
We recruited 150 US-based participants via Prolific and compensated them for their
time (50.7\% male, 49.3\% female; $M_{\text{age}} = 40.6$, $SD = 12.4$, range
18--81; 72.7\% White, 9.3\% Black, 9.3\% Mixed, 6.0\% Asian, 2.7\% Other),
randomly assigning each to one of the five conditions ($n = 30$ each).
As the test scenario was generated and evaluated in English, for the purposes of this experiment, non-English-speaking participants and those with technical issues preventing
completion were excluded.

\subsubsection{Procedure}
After providing informed consent, participants were randomly assigned to one of five experimental conditions. Participants in the two prebunking conditions first completed an AI Watchdog orientation introducing the five dark-pattern categories; those in the cognitive-forcing prebunking condition also completed a generative drafting exercise. Participants in the two just-in-time conditions received no advance briefing, and AI Watchdog instead intervened during the conversation when its classifier detected a dark pattern. Control participants received neither a pre-task orientation nor in-conversation intervention. All participants then completed the same two conversational tasks in a fixed order: a 15-turn trip-planning task followed by a 13-turn work-presentation task. In the just-in-time conditions, AI Watchdog continuously monitored the exchange. When a dark pattern was detected, the companion entered ``Barking Mode,'' displayed the suspected category, and presented a turn-specific reflection prompt. Rather than issuing a verdict or recommendation, the prompt asked participants to reconsider the exchange in their own terms. These reflection prompts were generated dynamically from the ongoing conversation rather than drawn from a fixed set, allowing them to respond to the specific preceding exchange. Both conditions used the same prompt generator and system instructions; they differed only in whether participants were required to respond. In the condition without cognitive forcing, participants could dismiss the prompt and continue immediately. In the cognitive-forcing condition, the conversation remained blocked until participants entered a free-text response. No minimum length, content requirement, or correctness criterion was imposed, so the intervention required participants to articulate a response before proceeding rather than merely acknowledge the warning with a click~\cite{buccinca2021trust}. After completing both tasks, participants completed a post-interaction survey measuring self-reported awareness of dark-pattern presence, trust in AI, and misinformation susceptibility. Demographic information was also collected. The full study took approximately 30--40 minutes to complete.

\begin{figure*}[t]
    \centering
    \includegraphics[width=\textwidth]{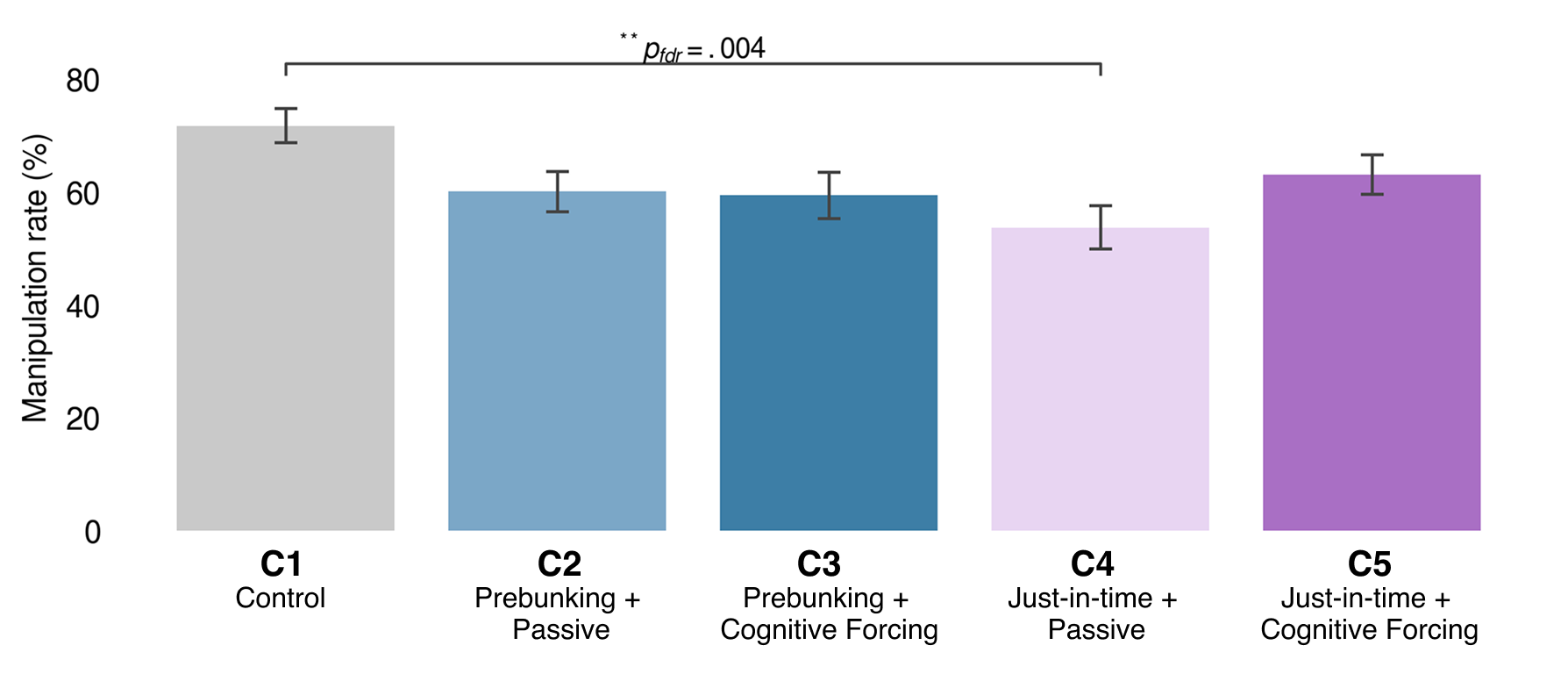}
    \caption{Rate of compliance with AI-steered decisions across five conditions: control; prebunking without cognitive forcing; prebunking with cognitive forcing; just-in-time without cognitive forcing; and just-in-time with cognitive forcing. Bars are mean compliance rates; error bars are standard errors. Kruskal--Wallis: $H(4) = 13.27$, $p = .010$. Only just-in-time without cognitive forcing differed significantly from control after Benjamini--Hochberg correction ($^{**}p_{\text{fdr}} = .004$).}
    \label{fig:manipulation}
\end{figure*}

\subsubsection{Ethics}
This research was reviewed by, and determined to be exempt by the MIT Committee on the Use of Humans as Experimental Subjects under Protocol \#E-7634.

\section{Results}
\label{sec:results}
We evaluated whether four intervention configurations of AI Watchdog, varying in nudge timing (prebunking vs.\ just-in-time) and engagement mode (without vs.\ with cognitive forcing), improved users' ability to recognize and resist conversational AI dark patterns. We assessed three complementary outcomes: explicit flagging of dark pattern turns, compliance with AI-steered recommendations including dark pattern, and post-task self-reported awareness.

For explicit flagging, no intervention significantly outperformed the control. Participants rarely flagged dark pattern turns across all conditions, with a median flagging rate of 0\% in every group and 80\% of participants flagging no manipulative turn at all. Mean flagging rates ranged from 0.7\% to 10.7\%, but the overall condition effect did not reach significance. Self-reported awareness showed a similar pattern, with no significant differences across conditions. Together, these results suggest that neither advance instruction nor real-time support reliably changed participants' explicit or retrospective recognition of dark patterns.

The behavioral outcome showed a different pattern. The just-in-time intervention without cognitive forcing was the only condition that significantly reduced compliance with AI-steered recommendations relative to control, lowering the compliance rate from 71.7\% to 53.7\%, an 18 percentage-point reduction ($p_{\text{fdr}}=.004$). This suggests that a timely, dismissable warning could influence users' subsequent decisions even when it did not measurably increase explicit flagging or self-reported awareness. By contrast, neither cognitive-forcing condition significantly reduced compliance relative to control. The just-in-time condition with cognitive forcing had a compliance rate of 63.0\% and did not differ significantly from either control ($p_{\text{fdr}}=.147$) or the corresponding just-in-time condition without cognitive forcing ($p_{\text{fdr}}=.194$), indicating no clear additional behavioral benefit from requiring a written response.

Classifier performance also did not differ significantly across experimental conditions, reducing the likelihood that uneven detection quality accounts for the observed differences in compliance. Exploratory analyses further suggest that recognition and resistance may vary across both users and dark-pattern types. Lower misinformation susceptibility was associated with higher flagging rates ($r=.190$, $p=.020$) but not with compliance, whereas higher post-task AI trust was associated with greater compliance ($r=.220$, $p=.007$) and lower self-reported awareness ($r=-.334$, $p<.001$). Pattern-level results showed a similar dissociation: sycophancy had the highest compliance rate (77.0\%) despite low flagging (4.0\%), while anthropomorphization was flagged most often (14.0\%) but still produced high compliance (73.3\%). These post-hoc findings reinforce the distinction between noticing potentially manipulative behavior and resisting its influence, while motivating further study of which users and conversational patterns may benefit most from defensive interventions.

These findings provide a more nuanced response to our preregistered hypotheses. \textbf{H1 was only partially supported.} The four intervention conditions did not consistently increase explicit flagging or reduce compliance relative to control. However, the just-in-time condition without cognitive forcing produced a significant reduction in compliance, indicating that at least one intervention configuration improved behavioral resistance even though recognition outcomes remained unchanged.

\textbf{H2 was not supported.} Conditions with cognitive forcing did not produce higher flagging rates or lower compliance rates than their corresponding conditions without cognitive forcing. In particular, adding a required written response to the just-in-time warning did not yield a statistically detectable advantage over the same warning without cognitive forcing. This suggests that increasing required engagement did not strengthen the intervention in the manner predicted.

\textbf{H3 was also not supported.} Contrary to our preregistered expectation that just-in-time delivery with cognitive forcing would produce the strongest defensive effect, the largest observed reduction in compliance occurred in the just-in-time condition without cognitive forcing. The cognitive-forcing variant did not significantly differ from either control or its corresponding non-forcing condition.

Taken together, these results refine our initial expectations about how defensive interfaces should operate in multi-turn AI conversations. The findings suggest that effectiveness may depend less on maximizing explicit recognition or deliberative effort and more on delivering timely, low-friction support at the moment users encounter potentially manipulative content. They also reinforce the importance of evaluating recognition and behavioral resistance as distinct outcomes, since an intervention may alter users' decisions without measurably increasing their explicit awareness of the dark pattern.

\subsection{No Intervention Increased Explicit dark pattern Flagging}

We report the preregistered primary outcome first. Flagging rates were low across all conditions, with a median of 0\% in every
group: 120 of 150 participants (80\%) flagged no manipulative turn at all. Means
ranged from $M = 0.7\%$ ($SD = 2.5\%$) in the just-in-time condition with cognitive
forcing to $M = 10.7\%$ ($SD = 18.6\%$) in the prebunking condition with cognitive forcing,
with both prebunking conditions (without cognitive forcing: $M = 10.0\%$, $SD = 19.8\%$;
with cognitive forcing: $M = 10.7\%$,
$SD = 18.6\%$) numerically above control ($M = 3.7\%$, $SD = 8.5\%$) and both
just-in-time conditions (without cognitive forcing: $M = 4.3\%$, $SD = 13.0\%$;
with cognitive forcing: $M = 0.7\%$,
$SD = 2.5\%$). The Kruskal--Wallis test was not significant,
$H(4) = 9.08$, $p = .059$, $\varepsilon^2 = .035$, and post-hoc Dunn tests yielded
no significant pairwise differences, though both prebunking conditions versus the
just-in-time condition with cognitive forcing
approached marginal significance ($p_{\text{fdr}} = .084$).

The just-in-time condition with cognitive forcing produced the numerically lowest rate
of any group, though no pairwise contrast
reached significance. 

\subsection{Just-in-Time Feedback Without Cognitive Forcing Reduces Compliance with AI-Steered Decisions}

We next examined the preregistered secondary outcome, compliance with scenario's AI-steered recommendations.
Compliance rates differed significantly across conditions, $H(4) = 13.27$,
$p = .010$, $\varepsilon^2 = .064$ (Kruskal--Wallis; see
Figure~\ref{fig:manipulation}). Control participants followed the
scenario's recommendation most frequently ($M = 71.7\%$, $SD = 16.6\%$, $Mdn = 75\%$),
while the just-in-time condition without cognitive forcing showed the lowest compliance rate
($M = 53.7\%$, $SD = 20.9\%$, $Mdn = 60\%$). Prebunking conditions were
intermediate (prebunking without cognitive forcing: $M = 60.0\%$, $SD = 19.5\%$;
prebunking with cognitive forcing: $M = 59.3\%$, $SD = 22.6\%$), as was the
just-in-time condition with cognitive forcing
($M = 63.0\%$, $SD = 19.0\%$).

Post-hoc Dunn tests with Benjamini--Hochberg correction showed that only the control
versus just-in-time-without-cognitive-forcing contrast was significant ($p_{\text{raw}} = .0004$, $p_{\text{fdr}} =
.004$), with 50\% of control participants
choosing the lower-rated, higher-priced hotel against 13\% in the just-in-time
condition without cognitive forcing. Control versus
prebunking was marginal (both $p_{\text{fdr}} = .073$), and the just-in-time condition
with cognitive forcing differed from
neither control ($p_{\text{fdr}} = .147$) nor its counterpart without cognitive
forcing ($p_{\text{fdr}} = .194$).
Just-in-time feedback without cognitive forcing was thus the only intervention to significantly
reduce the compliance rate relative to control. The same nudge with a
cognitive forcing function did not separate from control even marginally
($p_{\mathrm{fdr}} = .147$, against $p_{\mathrm{fdr}} = .073$ for both
prebunking conditions), though the two just-in-time conditions did not differ significantly from
each other. H1, which was conjunctive, was therefore not supported: no condition improved
flagging, and only just-in-time feedback without cognitive
forcing reduced the compliance rate. H2 and H3 were not supported: the cognitive forcing conditions did not significantly outperform their counterparts without cognitive forcing on either outcome, and the numerical ordering ran opposite to the predicted direction.

\subsection{Neither Prebunking nor Cognitive Forcing Increases Self-Reported Awareness of Dark Patterns in AI}

We assessed whether interventions elevated participants' subjective awareness of the presence of dark patterns. Self-reported awareness of the dark patterns did not differ significantly across conditions, $H(4) = 4.12$, $p = .391$. Mean scores clustered near the midpoint of the 1--7 scale across all groups, ranging from $M = 3.12$ ($SD = 1.51$, $Mdn = 2.70$) in the just-in-time condition without cognitive forcing to $M = 3.55$ ($SD = 1.24$, $Mdn = 3.80$) in the prebunking condition with cognitive forcing, and were comparable to the control ($M = 3.51$, $SD = 1.22$, $Mdn = 3.60$). Notably, the just-in-time condition without cognitive forcing reduced the compliance rate without a measurable increase in either explicit flagging or post-task self-reported awareness.

\subsection{Individual Differences in Flagging, Awareness, and Compliance}

The analyses in this subsection are post-hoc and exploratory. We examined how
misinformation susceptibility (MIST-20) and trust in AI related to key outcomes.
Both instruments were administered after the task, so both are post-treatment
measures: participants' responses could have been shaped by the conversation
itself or, in the four intervention conditions, by the intervention. Mean scores did not differ
significantly across conditions (MIST-20: $F(4, 145) = 0.92$, $p = .456$; AI
trust: $F(4, 145) = 0.59$, $p = .668$), but this does not establish that they
were unaffected. We therefore report these as associations measured after the
fact, not as pre-existing traits.

Participants scored relatively high on the MIST-20 ($M = 81.1\%$, $SD = 16.0\%$),
indicating comparatively low susceptibility to misinformation. Higher MIST-20
scores, that is lower susceptibility, were associated with a greater flagging
rate, $r = .190$, $p = .020$, but not with the compliance rate, $r = -.07$,
$p = .370$. Participants reported moderately high AI trust ($M = 4.57$,
$SD = 1.36$). Higher trust was associated with a greater compliance rate,
$r = .220$, $p = .007$, and with lower self-reported awareness, $r = -.334$,
$p < .001$. These are bivariate correlations; we did not fit a regression model,
so they describe association rather than prediction.

These associations do not displace condition, which produced the only
significant effect on the compliance rate. Their interest lies in flagging and
self-reported awareness, where no condition effect was detected and where the two
measures nonetheless varied systematically with outcomes.

\subsection{Exploratory Variation Across Pattern-Specific Items}
\label{sec:per-pattern}
The per-pattern analyses are post-hoc and exploratory. Sycophancy was the most behaviorally pervasive dark pattern with the highest compliance  ($77.0\%$) and low flagging ($4.0\%$), suggesting that flattering behavior was rarely recognized despite strongly shaping choices. Anthropomorphization was flagged most often ($14.0\%$) but still yielded high compliance ($73.3\%$), indicating that recognition did not reliably support resistance. Brand bias produced the lowest compliance ($29.3\%$), possibly because its binary, explicitly named alternatives made the framing easier to scrutinize. Sneaking and harmful generation were rarely flagged ($2.0\%$ each), with compliance rates of $68.0\%$ and $60.0\%$, respectively. However, harmful generation occurred at the final turn of both tasks, where 54 of 300 decision points ($18\%$) were unanswered and coded as non-compliance; its compliance estimate is therefore conservative, and both compliance and flagging may partly reflect attrition or fatigue rather than category-specific effects. These comparisons depend on the category validity established in Section~\ref{sec:scenario-eval}; accordingly, sneaking, which received the lowest coder confirmation, should be interpreted with particular caution.
Lower misinformation susceptibility was associated with higher flagging rates ($r = .190$) but not with compliance ($r = -.07$, n.s.). In contrast, higher trust in AI was associated with greater compliance with AI-steered decisions ($r = .220$) and lower self-reported awareness of manipulation ($r = -.334$). These associations indicate that the factors related to detecting dark patterns differ from those related to resisting their influence.

% An exploratory item-level pattern also emerged for brand bias. The scenario recommended Quora (100M affected users, described in emotionally rich detail) over Facebook (533M affected users, described in a single neutral sentence). Although the classifier detected brand bias in $81\%$ of turns, the just-in-time cognitive-forcing condition produced no participant flagging ($0\%$) and the highest compliance ($43\%$), compared with $23\%$ in control. One possibility is that mandatory reflection encouraged participants to rationalize rather than challenge the recommendation. However, with only 30 participants per condition for a single item and no corrected inferential test supporting this difference, this interpretation remains speculative. In contrast, prebunking with cognitive forcing matched control-level compliance ($23\%$) while producing the highest flagging rate for the item ($13\%$). Whether this apparent benefit of prior awareness generalizes beyond this scenario cannot be determined from the present data.

\section{Discussion}
\label{sec:discussion}
Our findings provide a nuanced response to our preregistered hypotheses. H1 was partially supported: just-in-time feedback without cognitive forcing significantly reduced compliance with manipulative AI recommendations, despite no increase in explicit flagging or post-task awareness. H2 and H3 were not supported: cognitive forcing provided no detectable benefit, and contrary to our prediction, the strongest defensive effect occurred in the just-in-time condition without cognitive forcing. The protective effect of cognitive forcing documented in single-shot decisions therefore did not appear to transfer to our multi-turn setting.

These findings align with prior work showing that cognitive forcing can fail against biased AI~\cite{vejandla2025impacts}, recognition of manipulative techniques may not improve decision quality~\cite{pennycook2024inoculation}, and resistance can occur without explicit recognition~\cite{tang2025dark}. Together, our results suggest that protective interventions should be tested directly in multi-turn interactions rather than assuming that effects from other decision contexts will transfer. With these findings in mind, we now turn to their broader implications and discussion.

\subsection{More Friction Did Not Yield More Defense}
Just-in-time warnings without cognitive forcing shifted behavior meaningfully; adding a deliberative prompt to the same nudge yielded no detectable benefit over control. This does not replicate, in a multi-turn setting, the pattern reported for single-shot AI-assisted decision-making, in which prompting reflection deepens the defensive effect~\cite{buccinca2021trust}. That assumption was already under
strain: partial explanations reduce overreliance less than full
ones~\cite{de2025cognitive}, and cognitive forcing has failed to mitigate
bias in AI-assisted decisions~\cite{vejandla2025impacts}.

What we observed is behavioral: the just-in-time condition without cognitive forcing differed from control, the one with cognitive forcing did not. We did not
measure cognitive load, attention, or deliberation, so the following is a
hypothesis rather than a finding. One possible explanation draws on dual-process
theory~\cite{kahneman2011thinking, evans2013dual}: warnings that require no response may operate
through fast heuristic processing, enough to shift behavior without demanding
sustained attention, whereas requiring a written reply may recruit slower
deliberation that competes for the attention needed to scrutinize the message.
The Elaboration Likelihood Model~\cite{petty1986elaboration} would predict this,
since users under load default to peripheral cues, here the manipulative framing.
What the results do support is narrower and still actionable: the low-friction
variant was the one that worked, so designers of dark pattern detection systems
have reason to test low-load interfaces before assuming that added friction
helps.

Brand bias, as a dark pattern, illustrates this sharply. Despite the classifier detecting it in the majority of cognitive forcing turns, participants in the just-in-time condition with cognitive forcing flagged no turns and showed the numerically highest compliance rate. One untested explanation is motivated reasoning ~\cite{kunda1990motivation, lord1979biased} in which forced introspection becomes rationalization; the item-level difference itself was not inferentially tested and should be read as exploratory. Both prebunking conditions produced the best
flagging rates, with the generative drafting exercise in the prebunking condition with cognitive forcing adding nothing over the
briefing alone in the prebunking condition without cognitive forcing, yet compliance rates remained at control levels, suggesting priming
may sharpen recognition without driving defense. Cognitive depth should therefore not be assumed to scale with defensive benefit.

\subsection{Individual Differences in Flagging, Awareness, and Compliance}
Lower misinformation susceptibility was associated with higher dark-pattern flagging ($r = .190$) but not with compliance ($r = -.07$, n.s.), while higher post-task AI trust was associated with greater compliance ($r = .220$) and lower self-reported awareness ($r = -.334$). This pattern suggests that recognizing a dark pattern and resisting its influence may depend on different user characteristics: being better at identifying potentially misleading content does not necessarily make a user less likely to follow an AI-steered recommendation. The association with AI trust is particularly important for defensive interface design. Participants reporting higher trust after the interaction were both more likely to comply with AI-steered decisions and less likely to report awareness of manipulation. Because trust was measured post-task, these findings should be interpreted as associations rather than evidence of a stable pre-existing vulnerability. Even so, they suggest that protective systems should not rely on users to accurately judge when they need additional safeguards. Designers and policymakers should also be cautious about overstating AI trustworthiness, since stronger trust may coincide with both greater receptivity to AI recommendations and reduced awareness of potentially manipulative behavior.

\subsection{Ethical Implications of Steering Behavior Through Defensive Design}

The compliance rate reduction produced by just-in-time feedback without cognitive forcing demonstrates
that a defensive interface can meaningfully redirect user behavior, and in that condition it did
so without participants reporting any greater awareness of manipulation. That
combination is what warrants caution. A monitor that reframes another system's
output while the user is reading it is, structurally, a second layer of persuasion
operating on the first, and the mechanism that defends users here could be turned to
steering them if deployed by a party with misaligned incentives. This concern compounds with a known property of persuasion: users under load
default to peripheral cues~\cite{petty1986elaboration}, and in conversational AI
the system generating the persuasive content also controls the pace, framing, and
cognitive demands of the exchange. As such, independent auditing of the monitor, not only of the system it monitors, should be treated as a baseline requirement.

Secondly, local inference is a must: the AI Watchdog should run on the user's own
device rather than sending conversations to another company. A monitor of this kind
must read every turn, and 82\% of users rate chatbot conversations as sensitive or
highly sensitive, above email and social media posts~\cite{tran2025privacynorms}.
The risk is compounded by what such a monitor could infer. Our results show that
users higher in AI trust are both more manipulable and less aware of it, so a
system that observed conversations at scale could identify which users are easiest
to steer. That profile is precisely what a platform with misaligned incentives
would want, which is why the tool that detects manipulation should not also be collecting data that leaves users vulnerable.

\subsection{Design Implications for Defensive Conversational AI Systems}
Our findings suggest several directions for the design of defensive interfaces, while also pointing to questions that require further testing. First, the just-in-time condition without cognitive forcing was the only intervention that significantly reduced compliance relative to control, whereas requiring a written response did not provide a detectable additional benefit. This suggests that low-friction support delivered at the moment of potential manipulation may be worth prioritizing before adding more demanding forms of engagement. Second, the numerical pattern across conditions raises the possibility that awareness-building and behavioral support may play different roles: prebunking conditions produced higher flagging rates, while just-in-time feedback without cognitive forcing produced the strongest behavioral effect. Future systems could therefore explore combining advance preparation with lightweight in-the-moment cues, consistent with inoculation theory's emphasis on prior exposure~\cite{roozenbeek2020prebunking}. Finally, because explicit flagging remained low even when compliance changed, defensive interfaces may benefit from evaluating success not only through whether users explicitly identify manipulation, but also through whether the intervention helps them make more resistant decisions. Taken together, these results motivate further investigation of timely, low-friction interventions that support behavioral resistance without requiring users to continuously identify or deliberate over every potentially manipulative turn.

\subsection{Limitations and Future Work}
\label{sec:limitations}

\textit{Classifier evaluation.} The classifier was chosen as the best of nine
candidates on a 90-turn selection set, where it agreed with the scripted
positions 92\% of the time; during the study itself, agreement was closer to 80\%
(Section~\ref{sec:apparatus}). Selection-set figures can therefore overstate
in-use performance, and systems of this kind should report agreement
measured during deployment, not only at model-selection time. 
Future work should also evaluate the classifier against human-coded ground truth, including whether the category it displays is the correct one. 

\textit{Statistical power.} With $n = 30$ per condition, the experiment establishes a
starting point for understanding how nudge timing (prebunking vs.\ just-in-time) and engagement mode (without vs.\ with cognitive forcing) affect
detection of and defense against dark patterns in multi-turn AI conversations.
However, the design reliably detects medium-sized differences and not the small
effects typical of misinformation correction research. The control versus
just-in-time-without-cognitive-forcing contrast survived
correction for multiple comparisons and is the finding we treat as established; the
marginal control-versus-prebunking contrasts ($p_{\text{fdr}} = .073$) are precisely
those a larger sample might resolve in either direction. Our null results for the
flagging rate and self-reported awareness should therefore be read as an
absence of large effects rather than evidence of no effect.

\textit{Language and population.} The test scenario was generated, and the
classifier's definitional prompt written, in English, and we excluded
non-English-speaking participants. Two of the five patterns are defined partly by
linguistic register rather than propositional content: sycophancy by hollow
flattery, anthropomorphization by first-person claims to feeling or memory. Whether
these cues survive translation is not obvious, since norms around deference,
praise, and indirectness differ across languages and cultures, and a turn that reads
as manipulative flattery in American English may read as ordinary politeness
elsewhere. Our US-based sample also skewed White (72.7\%), and the decisions themselves are culturally situated, with the trip-planning task assuming
familiarity with Western travel conventions and star-rating systems. Future work should include testing out the effectiveness of the AI Watchdog in multi-cultural and multi-lingual contexts.

\textit{Ecological validity.} The experimental scenario was a mock-up, where participants were asked to plan for a trip and for a work presentation, with each dark pattern presented at a predetermined turn index and followed immediately by a decision point. Consequently, following the AI's recommendation entailed no real financial or social consequences. Participants did not actually overpay for the lower-rated hotel, experience the consequences of staying there, or defend the framing of their selected case study in their presentation to real colleagues. The absence of such consequences may have reduced the incentive to scrutinize recommendations as carefully as they would in real-world settings. Deploying AI Watchdog as a browser extension during participants' own AI-assisted interactions would address this limitation in two ways. First, interactions would be unscripted, allowing the classifier to be evaluated on naturally occurring instances of manipulation rather than experimentally injected ones. Second, decisions would carry real-world financial or social consequences, providing a more ecologically valid measure of whether participants comply with AI-generated recommendations.

\section{Conclusion}

As conversational AI increasingly influences decisions across everyday and consequential contexts, users need practical support for recognizing and resisting manipulative interaction patterns. This work examined that challenge through \textit{AI Watchdog}, a proactive monitoring interface that detects conversational dark patterns and surfaces interventions during live, multi-turn interaction.

Across a preregistered five-condition experiment with 150 participants, we found that explicit recognition and behavioral resistance did not move together. Participants rarely flagged dark-pattern turns, and self-reported awareness did not significantly differ across conditions. Yet the just-in-time intervention without cognitive forcing significantly reduced compliance with AI-steered recommendations from 71.7\% in the control condition to 53.7\%, an 18 percentage-point reduction. Requiring an additional written response did not provide a detectable advantage over the same just-in-time intervention without cognitive forcing. These findings suggest that effective defensive support may not require users to explicitly identify every manipulative turn or engage in additional deliberative effort.

Our exploratory analyses further reinforce this distinction. Lower misinformation susceptibility was associated with greater flagging but not lower compliance, while higher post-task AI trust was associated with greater compliance and lower reported awareness. Pattern-level variation showed a similar separation: sycophancy produced the highest compliance despite low flagging, while anthropomorphization was more frequently recognized but still associated with high compliance. Together, these findings indicate that noticing manipulation and resisting its influence are related but distinct challenges that should be evaluated separately.

More broadly, this study provides an initial empirical foundation for designing user-facing defenses against conversational AI dark patterns. AI Watchdog demonstrates a pathway toward independent monitoring, while our multi-turn test environment provides a reusable setting for studying how manipulative behaviors affect human decisions and how interventions might mitigate them. Although further work is needed across larger populations, conversational contexts, and privacy-preserving deployments, our results point to a promising direction: timely, low-friction interventions may help users maintain greater autonomy in increasingly persuasive AI interactions.

\section{Generative AI Disclosure}
Generative AI was used to refine phrasing in author-drafted text. GPT-4o powered the AI-generated test scenario, and Llama-3.3-70B-Instruct served as the classifier for detecting dark patterns in messages. Visual materials for the companion were generated using Google's Nano Banana.

\bibliographystyle{ACM-Reference-Format}
\bibliography{references}

\end{document}